\documentclass[conference]{IEEEtran}
\IEEEoverridecommandlockouts
\usepackage{cite}
\usepackage{amsmath,amssymb,amsfonts}
\usepackage{algorithmic}
\usepackage{graphicx}
\usepackage{textcomp}
\usepackage{xcolor}

\usepackage{tabularx}
\def\BibTeX{{\rm B\kern-.05em{\sc i\kern-.025em b}\kern-.08em
    T\kern-.1667em\lower.7ex\hbox{E}\kern-.125emX}}
\newcommand{\coo}{\ensuremath{\mathrm{CO_2}}}

\usepackage[]{footmisc}
\usepackage{comment}

\begin{document}

\title{Evaluating Robustness of Reinforcement Learning Algorithms for Autonomous Shipping\\
\thanks{This work is conducted within the DEFRA AHOI project, funded by the Belgian Royal Higher Institute for Defence, under contract number 23DEFRA002.
}
}  

\author{\IEEEauthorblockN{Bavo Lesy*}
\IEEEauthorblockA{\textit{Faculty of Applied Engineering, IDLab} \\
\textit{University of Antwerp - imec}\\
Antwerp, Belgium \\
bavo.lesy@uantwerpen.be}
*Corresponding author

\and
\IEEEauthorblockN{Ali Anwar (Member IEEE)}
\IEEEauthorblockA{\textit{Faculty of Applied Engineering, IDLab} \\
\textit{University of Antwerp - imec}\\
Antwerp, Belgium \\
ali.anwar@uantwerpen.be}

\and
\IEEEauthorblockN{Siegfried Mercelis}
\IEEEauthorblockA{\textit{Faculty of Applied Engineering, IDLab} \\
\textit{University of Antwerp - imec}\\
Antwerp, Belgium \\
siegfried.mercelis@uantwerpen.be}
}

\maketitle

\begin{abstract} 
Recently, there has been growing interest in autonomous shipping due to its potential to improve maritime efficiency and safety.
The use of advanced technologies, such as artificial intelligence, can address the current navigational and operational challenges in autonomous shipping. In particular, inland waterway transport (IWT) presents a unique set of challenges, such as crowded waterways and variable environmental conditions. In such dynamic settings, the reliability and robustness of autonomous shipping solutions are critical factors for ensuring safe operations. This paper examines the robustness of benchmark deep reinforcement learning (RL) algorithms, implemented for IWT within an autonomous shipping simulator, and their ability to generate effective motion planning policies.
We demonstrate that a model-free approach can achieve an adequate policy in the simulator, successfully navigating port environments never encountered during training. We focus particularly on Soft-Actor Critic (SAC), which we show to be inherently more robust to environmental disturbances compared to MuZero, a state-of-the-art model-based RL algorithm. In this paper, we take a significant step towards developing robust, applied RL frameworks that can be generalized to various vessel types and navigate complex port- and inland environments and scenarios.
\end{abstract}


\section{Introduction}
In recent years, the field of autonomous navigation and its logistic applicability have garnered significant attention, with promising results in industrial settings such as warehousing \cite{FRAGAPANE2021405} and airborne deliveries \cite{drones8060220}. However, an equally promising, yet less practiced area is autonomous shipping, with potential especially in autonomous inland waterway transport (IWT). IWT is one of the most \coo-efficient transport modes per ton of goods carried \cite{doi/10.2832/03796}, making it a suitable candidate to reduce emissions. However, IWT has been declining for several years in countries such as The Netherlands and Belgium due to a shortage of skippers and the harsh competition of road transport. Autonomous shipping can greatly benefit IWT by reducing the number of skippers required, consequently reducing the operating costs and improving the competitiveness of IWT. The Central Commission for the Navigation of the Rhine (CCNR) has defined six levels of autonomy for autonomous vessels (AVs)\footnote{https://www.ccnr.eu/12050000-en.html}. Currently, most autonomous vessels reach level 2 autonomy, still requiring a skipper or human operator to be involved at all times to take control when necessary. In order to reduce the number of skippers, at least level 3 autonomy should be achieved so that a single human operator can supervise multiple vessels at the same time.  

Much research aims at improving vessel autonomy, with major advancements in motion planning methods that utilize supervised deep learning techniques \cite{9756903}. Motion planning deals with finding an optimal and feasible route for a (semi-)autonomous vehicle to reach its goal, whilst avoiding collisions and adhering to various constraints. Within the field of motion planning, deep reinforcement learning (RL) has shown promising results, outperforming traditional control schemes \cite{9454561}\cite{Herremans_2023}. Traditional control methods for motion planning and collision avoidance, such as the dynamic window approach (DWA) \cite{580977} and model predictive control (MPC) \cite{8768044} require extensive parameter tuning to achieve optimal performance. Such methods are often tailored to specific scenarios, relying on predefined models of the vessel and its environment. Changes in the operational context, such as engine wear and tear, changes in climate conditions, or increased traffic in the area, may require the control system to be re-calibrated or re-designed.

RL, on the other hand, excels in such environments because of its data-driven nature. RL does not require predefined models or extensive manual tuning. In RL, an agent (or controller) aims to solve a sequential decision-making problem by learning through trial and error. The agent learns a policy $\pi$ that determines which actions to take in which state. Unlike supervised learning techniques, where explicit data labeling is required, RL techniques do not need a human-labeled dataset, instead learning from direct interaction with the environment, allowing it to adapt to a wide variety of conditions.The agent needs to have significant interaction with the environment before learning an adequate policy. Because of this, the environment in which the agent learns is typically a simulator in which the agent can explore freely, without the risks, time, or costs associated with performing a wrong action. However, learning a policy in a simulator and then transferring it to the real world has limited success, often due to the limitations and modeling errors of the simulator \cite{pmlr-v78-rusu17a}. Furthermore, these variations between the training and testing environment are often not taken into account. This is one of the key reasons why RL methods have not seen the same adoption in real-world applications as their supervised learning counterparts.

The field of Robust RL aims to bridge the gap between simulation and real life (sim2real) by accounting for these model uncertainties and the variability of the environment. Research in robust RL has shown improved success in closing the sim2real gap \cite{Peng_2018} by creating policies robust to these uncertainties. In the context of autonomous shipping in IWT, robust RL can play a key role in improving the operational use of these algorithms. Consider an autonomous barge, on course to pick up a large amount of cargo. When the barge gets loaded, its properties - and thus the required control actions - change. Can we learn a policy that is robust to this change? Or in an even broader context, can a policy be transferred between different vessels? By treating these changes as modeling errors or external disturbances in the simulation, we can address this question within the context of robust RL. Learning a policy robust to modeling errors and disturbances should also make the policy robust to changes in vessel and environmental conditions, thus improving the usability of RL methods and taking a significant step toward level 3 autonomy for AVs. In this work, we examine the robustness of benchmark RL algorithms against various changes in vessel and environmental conditions. 

The rest of this paper is structured as follows. Section \ref{sec:related works} covers related research in the fields of autonomous shipping and (robust) RL and its applications. Our methodology and main research contributions are explained in detail in Section \ref{sec:methodology}. The results are then presented and discussed in Section \ref{sec:results}, followed by a conclusion summarizing our findings and suggesting directions for future research in Section \ref{sec:conclusion}.

\section{Related Works}
\label{sec:related works}
\subsection{Autonomous Shipping}
Since the literature in the field of autonomous shipping is vast, we focus specifically on motion planning and collision avoidance. For a more extensive view of deep learning methods for autonomous vessels, the reader is referred to \cite{10021250} and \cite{Vagale2021}.
Zhang et al. \cite{jmse9070761} employ a hybrid motion planning approach, autonomously generating both global and local paths, based on the artificial potential field (APF) and velocity obstacle (VO) algorithms, respectively. Tang et al. \cite{9815528} utilize an LSTM model to predict the movement of AVs and plan a trajectory with MPC. Furthermore, Chun et al. \cite{CHUN2021109216} use RL to avoid collisions while complying with regulations. However, the assumption is made that all vessels share their position and velocity. In IWT, this is problematic since there are many static obstacles and smaller vessels that do not share this information. This is why in our work we rely on the observations of a ranging sensor to detect static and dynamic obstacles. 

Vanneste et al. \cite{9968678} focus on safe collision avoidance with static and dynamic obstacles, while proposing a model-predictive RL method for motion planning in a novel autonomous shipping simulator within the MOOS-IVP framework \cite{moos-ivp}. Recent work by Herremans et al. \cite{Herremans_2023} extends on this, using a model-based reinforcement learning (MBRL) approach for path planning and collision avoidance in the same simulator. Waltz et al. \cite{waltz20242levelreinforcementlearningships} propose 2-level RL, focusing on high-level path planning and low-level path following. However, these works focus solely on performance in one specific environment, not investigating the robustness of their solutions under varying environmental conditions.


\subsection{Robust RL}
Much research aims to improve the robustness of RL algorithms, with methods such as domain randomization \cite{tobin2017domainrandomizationtransferringdeep} and dynamics randomization \cite{Peng_2018} showing promising results and successfully transferring policies to the real world.  Zhang et al. \cite{pmlr-v202-zhang23bc} enhance robustness by taking into account disturbances in context transitions, while Zhang et al. \cite{NEURIPS2020_77441296} investigate robustness in a multi-agent setting. 

Pinto et al. \cite{pmlr-v70-pinto17a} introduce an adversarial learning method to increase policy robustness. A second agent is introduced, which tries to degrade the performance of the primary agent by introducing disturbances into the simulator, turning the problem into a two-player zero-sum game. This adversarial learning approach has been extended by Zhai et al. \cite{Zhai_Luo_Dong_Zhang_Wang_Yang_2022}, who utilize concepts of $H_\infty$-control to improve learning stability and robustness. Recent work by Sung et al. \cite{sung2024robustmodelbasedreinforcement} also addresses robustness from a control-theoretic perspective by incorporating $\mathcal {L}_1$ Adaptive Control into MBRL. Herremans et al. \cite{herremans2024robustmodelbasedreinforcementlearning} adapt MBRL to account for model uncertainties, significantly improving robustness. Rigter et al. \cite{NEURIPS2022_6691c5e4} introduce adversarial learning in the context of offline RL, increasing robustness in areas not covered by the initial dataset. Most of the existing literature focuses on introducing novel robustness methods whilst limiting the application to relatively simple environments.
In a more practical context, Wang et al. \cite{WANG2024107728} propose an adversarial RL approach to increase the robustness of AVs, focusing on the lower-level depth control of an autonomous underwater vehicle. Our work focuses on evaluating the robustness of RL methods for higher-level motion planning and collision avoidance on inland waterways within the MOOS-IVP autonomous shipping simulator \cite{9968678}.

\section{Methodology}
\label{sec:methodology}
\subsection{Simulation}
\label{subsec:simulation}
Since we focus on higher-level motion planning and collision avoidance, and not lower-level control, we employ a three-degree-of-freedom (3-DOF) kinematic model \cite{Herremans_2023} to simulate vessel movement. Using this model, the motion of a vessel at timestep $t$ can be represented by a pose vector $\mathbf{p_t}$ and a velocity vector $\mathbf{v_t}$, where: $\mathbf{p_t} = \begin{bmatrix} x_t, \ y_t, \ \theta_t \end{bmatrix}^T$, with $(x_t, y_t)$ representing the position of the vessel at time $t$ and $\theta_t$ the orientation of the vessel. $\mathbf{v_t} = \begin{bmatrix} s_t, \ \omega_t  \end{bmatrix}^T$ is used to describe the rate at which the position and orientation of the vessel change. $s_t$ denotes the linear velocity and $\omega_t$ the angular velocity of the vessel. The rudder (yaw) and thrust (surge) of the vessel can be controlled with input $\mathbf{u_t} = \begin{bmatrix} u_{thrust}, \ u_{rudder} \end{bmatrix}^T$. 
During simulation, $\mathbf{v_t}$ and $\mathbf{p_t}$ are updated as follows:
\begin{equation}\label{eq:acc}
a = \frac{u_{thrust}}{m}
\end{equation}
\begin{equation}\label{vt}
\mathbf{v_{t+1}} = \mathbf{v_t} + \begin{bmatrix} a \cdot dt \\  u_{rudder} \cdot R \cdot dt \end{bmatrix}
\end{equation}
\begin{equation}\label{pt}
\mathbf{p_{t+1}} = \mathbf{p_t} + \begin{bmatrix} \sin{\theta} (s \cdot dt) \\ \cos{\theta} (s \cdot dt) \\ \omega \cdot dt \end{bmatrix},
\end{equation}
where m is the mass of the vessel and R is the turn rate of the vessel. 

\subsection{Reinforcement Learning}\label{sub:rl}
In RL, an agent learns to solve a sequential decision-making problem through continuous interaction with an environment. In this work, we solve the problem of motion planning in the MOOS-IVP simulator. This problem can be mathematically formulated as a Markov Decision Process (MDP). In this framework, the problem can be defined by a tuple \((S, A, T, R, \gamma)\), where $S$ is the set of all possible states, $A$ is the set of all possible actions, $T$ represents the transition model and $R$ the reward model, which defines the return after taking an action in a given state. Lastly, $\gamma$ is the discount factor, which adjusts the weight of immediate and future return. The agent determines which action $a$ to be executed at each timestep $t$ by employing a policy function $\pi$. The objective (Eq. \ref{obj}) in RL is finding an optimal policy $\pi^*$ which maximizes the expected accumulated return across an episode:
\begin{equation}\label{obj}
J(\pi) = \mathbb{E}\left[ \sum_{t=0}^{H} \gamma^t R_t \mid \pi \right],
\end{equation}
where $H$ denotes the horizon or length of the episode. The optimal policy can then be defined as follows:
\begin{equation}\label{optimal}
\pi^*(s) = \arg\max_\pi \; J(\pi) 
\end{equation}

More formally, the problem of motion planning can be defined as follows: Given a starting location and a goal, determine the optimal path to the goal while avoiding obstacles. In each episode, the vessel starts at a random location within a randomized port environment, which includes static and dynamic obstacles, as well as a randomly placed goal. The optimal policy $\pi^*$ (Eq. \ref{optimal}) identifies the optimal path to the goal while avoiding obstacles by controlling the thrust and rudder of the vessel. However, the vessel does not know the location of obstacles or the port walls. Instead, it relies on a ranging sensor to observe its surroundings and detect obstacles. An episode ends when the goal is reached (positive return) or if a collision occurs (negative return). 

\subsection{Experimental Setup}\label{sub:setup} 
The goal of this work is to examine the robustness of RL algorithms. The algorithms used are the model-free Soft-Actor Critic (SAC) algorithm \cite{haarnoja2018soft} and the model-based MuZero algorithm \cite{Schrittwieser_2020}. MBRL combines traditional RL with planning by learning an internal transition model which approximates the dynamics of the actual environment. This model allows for future reasoning and planning on which actions to take. This has shown to reduce the number of environment interactions needed to achieve performance comparable to that of model-free RL \cite{Luo2024}. We choose to examine the robustness of SAC, as Eysenbach et al. \cite{eysenbach2022maximum} theoretically prove that the SAC algorithm is inherently robust to context disturbances. SAC is compared to MuZero, an algorithm with no inherent robust properties, as we reproduce the experiment by Herremans et al. \cite{Herremans_2023}. As seen in Eq. \ref{eq:acc} and Eq. \ref{vt}, the mass and turn rate of the vessel affect the kinematics of the simulation. To examine the robustness of the algorithms, we can vary these parameters to emulate vessels different from the training vessel and see how they perform. We adapted the simulator to allow for changes to these parameters during runtime. 

\begin{figure}[ht]
\centering
  \includegraphics[width=0.7\linewidth]{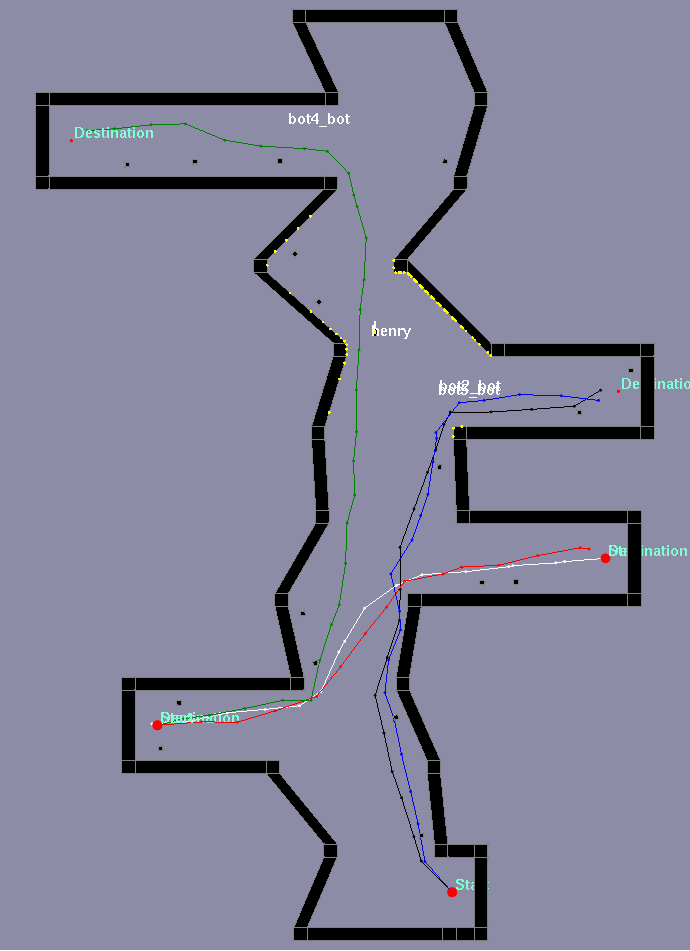}
  \caption{Port environment in the MOOS-IVP simulator. The various colored lines represent the routes of the dynamic obstacles, with the yellow dots indicating the obstacles detected with the ranging sensor.}
  \label{fig:sim}
\end{figure}

To train RL algorithms for this task, we use the custom MOOS-IVP simulator (Fig. \ref{fig:sim}) with the RLLib framework \cite{liang2018rllib} to implement distributed RL training across 20 parallel simulations.
\begin{figure*}[ht]
  \centering
  \includegraphics[width=0.45\linewidth]{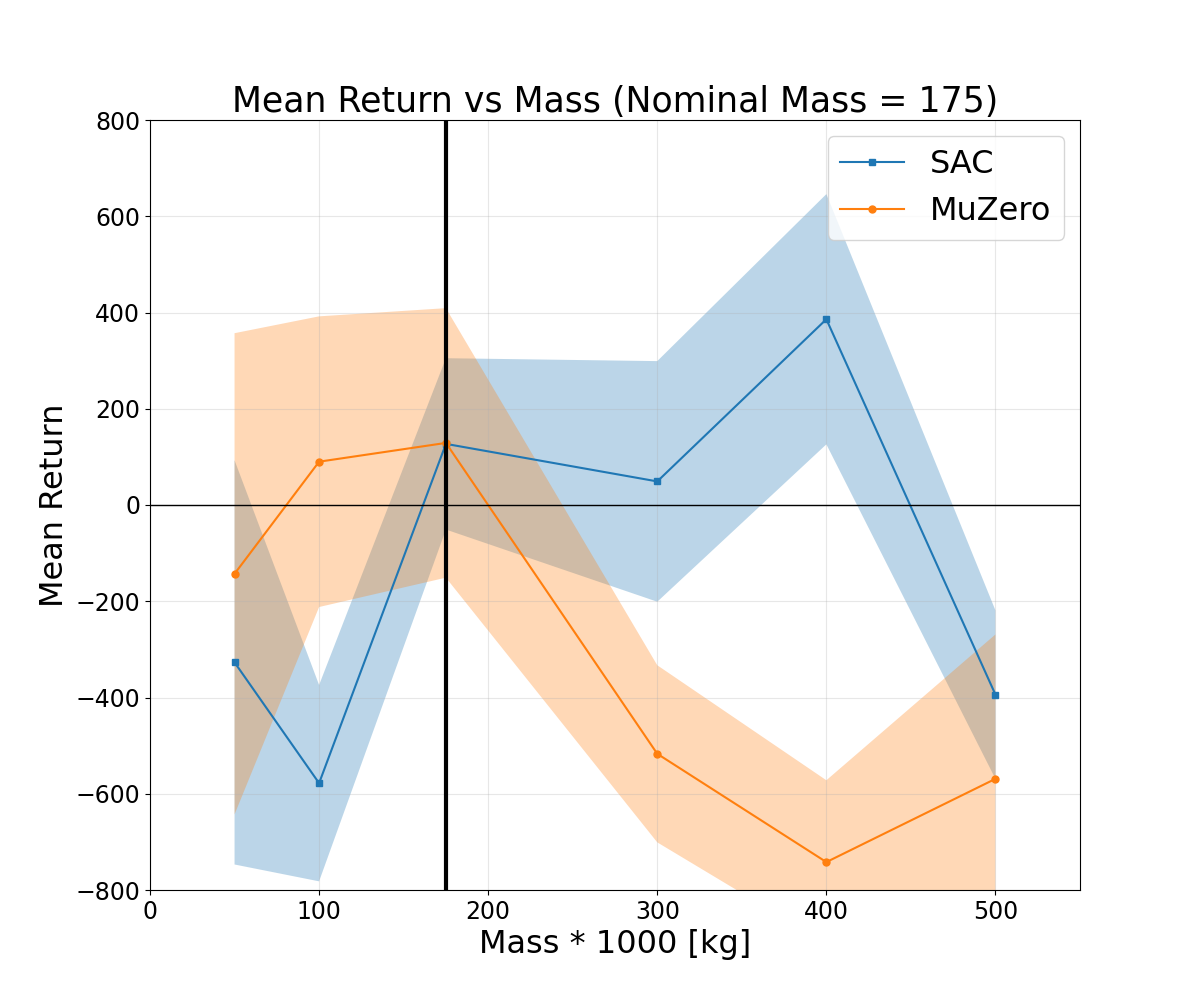}
  \includegraphics[width=0.45\linewidth]{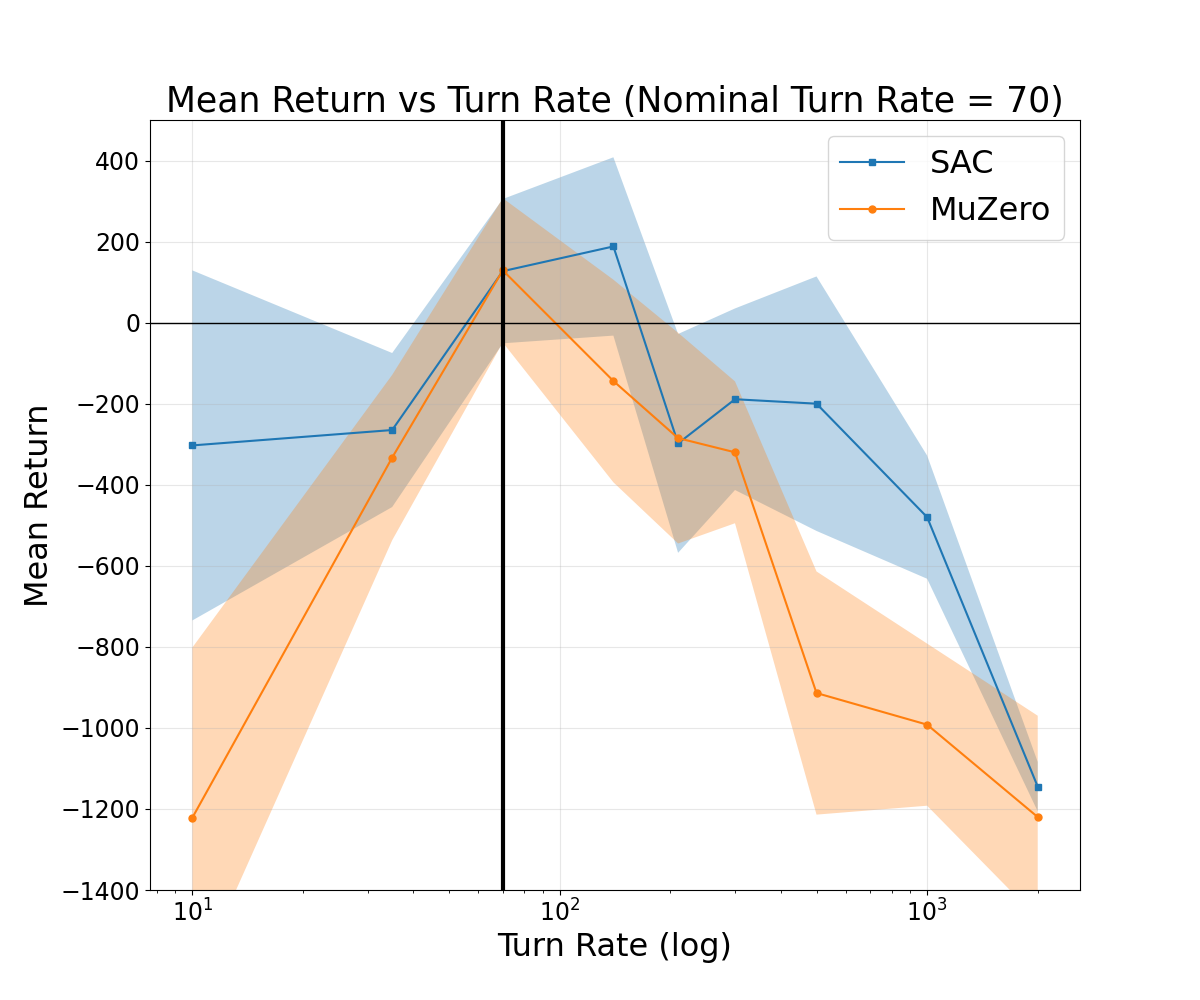}
  \caption{Mean return evaluation of SAC and MuZero with varying mass and turn rate values. The shaded regions indicate the standard deviation and the bold vertical lines indicate the nominal mass and turn rate of 175 000 kg and 70 respectively. The turn rate graph has a logarithmic scale on the x-axis.}
  \label{fig:robust}
\end{figure*}
\begin{figure}[ht]
  \centering
  \includegraphics[width=0.8\linewidth]{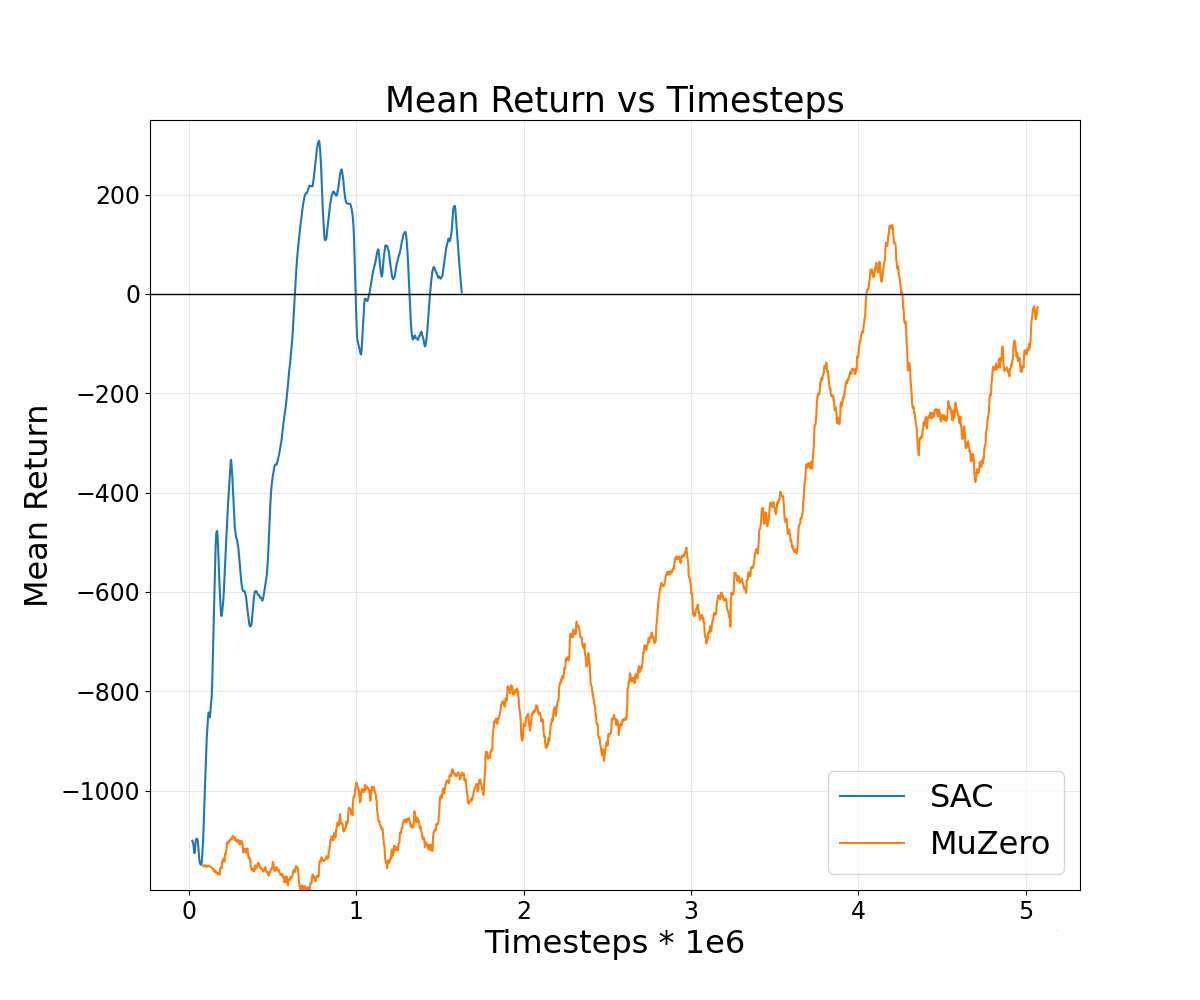}
  \caption{Mean return during the training process of SAC and MuZero in MOOS-IVP simulator.}
  \label{fig:mean_reward_1}
\end{figure}
\section{Results}
\label{sec:results}
Fig. \ref{fig:mean_reward_1} shows the mean return during the training process. A higher return indicates longer and more successful episodes with fewer collisions.  After training, both SAC and MuZero achieve similar performance. However, SAC achieves this performance after significantly fewer training timesteps. This could be because the model-free approach does not need to learn a transition model of the complex environment. In contrast to Herremans et al. \cite{Herremans_2023}, we show that a model-free approach can also achieve an adequate policy for motion planning in the MOOS-IVP simulator. To assess robustness, we modified the vessel mass $m$ and turn rate $R$ after training and analyzed the return compared to the standard values for both SAC and MuZero. On Fig. \ref{fig:robust}, we notice that SAC is more robust than MuZero. As the mass of the vessel increases, MuZero's performance rapidly deteriorates, whereas SAC maintains similar performance for larger mass values and degrades later. This can be attributed to the fact that SAC is based on maximum entropy RL \cite{haarnoja2018soft}, where an agent aims to maximize the expected return whilst acting as randomly as possible. Random actions can be beneficial in the presence of environmental disturbances, since previously learned actions may no longer be effective. At first glance, we would expect that SAC also performs better for the lower mass values. However, lowering the mass makes it easier to control the vessel, as the influence of the control action increases (Eq. \ref{eq:acc}). This could be why MuZero performs better in this case. 
We see similar results for the turn rate, with SAC generally performing substantially better for both higher and lower turn rates. 
Performance drops off only at relatively high turn rates, with decent return even at ten times the nominal turn rate. This could be because the vessel usually makes small turns, not significantly affected by the turn rate (Eq. \ref{vt}). However, lower turn rates degrade the performance faster, due to limiting the vessel's ability to make rapid turns when necessary.

\section{Conclusion and Future Work}
\label{sec:conclusion}
This paper examines the robustness of RL algorithms in a practical and complex environment. We extend previous work \cite{Herremans_2023} and show that a model-free approach can also successfully navigate various port and inland environments in less training time compared to a model-based approach. We show that SAC is indeed inherently more robust to environmental disturbances than MuZero. 
However, we only examined the mass and turn rate separately. Future work will include the incorporation of a more sophisticated kinematics model, which includes drag. Adapting the drag coefficient and drag area, along with the mass and turn rate, can emulate vessels with completely different shapes and sizes. 
Furthermore, we will evaluate existing robust RL methods \cite{herremans2024robustmodelbasedreinforcementlearning} \cite{NEURIPS2022_6691c5e4} in a more practical environment. Since we now have the ability to adapt the simulation parameters at runtime, we can implement adversarial learning methods, as those discussed in \cite{pmlr-v70-pinto17a} and \cite{Zhai_Luo_Dong_Zhang_Wang_Yang_2022}. 

Lastly, because robust RL aims to close the sim2real gap, we would like to transfer these approaches to a real-world unmanned surface vehicle (USV) and examine their effectiveness. For this, a number of implementation challenges have to be addressed, including installing an embedded device such as Nvidia Jetson to run the algorithm on the vessel. Furthermore, sensor data must be captured in a similar format as the simulated data, and the output actions to control the vessel's rudder and engine must be translated correctly.


\bibliographystyle{IEEEtran}
\bibliography{references}

\end{document}